\documentclass{article}

\PassOptionsToPackage{numbers, compress}{natbib}

\usepackage[final]{nips_2018}

\usepackage[utf8]{inputenc} % allow utf-8 input
\usepackage[T1]{fontenc}    % use 8-bit T1 fonts
\usepackage{hyperref}       % hyperlinks
\usepackage{url}            % simple URL typesetting
\usepackage{booktabs}       % professional-quality tables
\usepackage{amsfonts}       % blackboard math symbols
\usepackage{nicefrac}       % compact symbols for 1/2
\usepackage{microtype}      % microtypography

\usepackage{graphicx}
\usepackage{subcaption}
\usepackage{xcolor}
\usepackage{comment}
\usepackage{float}
\usepackage{enumitem}
\usepackage{lmodern}
\usepackage[ruled,vlined,linesnumbered]{algorithm2e}
\usepackage[title]{appendix}

\title{Weakly Supervised Active Learning with Cluster Annotation}

\author{
  Fábio Perez\thanks{The author was an intern at EPFL during the development of this work.} \\
  University of Campinas, Brazil\\
  \texttt{fabiop@dca.fee.unicamp.br} \\
  \And
  Rémi Lebret \phantom{abcde} Karl Aberer \\
  EPFL, Lausanne, Switzerland \\
  \texttt{\{remi.lebret,karl.aberer\}@epfl.ch} \\
}

\begin{document}

\maketitle

\vspace{0.3cm}
\begin{abstract}
In this work, we introduce a novel framework that employs cluster annotation to boost active learning by reducing the number of human interactions required to train deep neural networks. Instead of annotating single samples individually, humans can also label clusters, producing a higher number of annotated samples with the cost of a small label error. Our experiments show that the proposed framework requires 82\% and 87\% less human interactions for CIFAR-10 and EuroSAT datasets respectively when compared with the fully-supervised training while maintaining similar performance on the test set.
\end{abstract}
\vspace{0.3cm}

\section{Introduction}

Deep learning techniques represent the state-of-the-art in many machine learning problems in diverse fields such as computer vision, speech recognition, and natural language processing. However, training deep learning models usually require considerably large amounts of labeled data, which may be expensive to obtain.

Supervised learning is an effective way of training machine learning models, but requires the annotation of every sample in the training dataset. To reduce the need for labeled data, machine learning models can work under other supervision schemes. Unsupervised learning methods use unlabeled data during training. Semi-supervised learning methods can learn from both labeled and unlabeled data. Weakly supervised learning methods use data samples that may contain annotation noise (e.g., images annotated with Instagram hashtags~\cite{DBLP:conf/eccv/MahajanGRHPLBM18}).

Active learning is a subclass of semi-supervised learning methods in which an expert iteratively annotates data in order to maximize the model knowledge while minimizing the number of data labeling. A commonly used query strategy is the pool-based sampling, in which the active learning framework selects the most informative samples (i.e., samples that are expected to most increase the model knowledge) from a pool of unlabeled data based on some criteria.

We propose a novel framework that employs clustering to boost active learning by reducing the number of human interactions required to annotate datasets. Our method combines semi-supervised learning and weakly supervised learning. Instead of only annotating single samples, experts annotate clusters that have class consistency (i.e., the vast majority of samples belonging to the same class), significantly reducing the required number of human interactions to train a model.

\section{Related Work}

Despite being used in machine learning for a long time, it was only recently that active learning was applied to deep learning architectures. Most active learning methods for convolutional neural networks (CNNs) are based on uncertainty measurements. However, it is hard to measure uncertainty in deep learning models \cite{what_unc}.

\citet{ceal} apply active learning to CNNs for image classification. To select the most uncertain samples, they use the output of the softmax layer after inputting an unlabeled image to the network. One of the criteria used for selecting samples is the maximum entropy of the softmax layer. 

\citet{dropout_al} use Monte Carlo dropout, introduced in \citet{DBLP:conf/icml/GalG16}, to measure Bayesian uncertainty in CNNs for selecting samples for active learning. They show that collecting multiple outputs from networks with dropout during inference results in better uncertainty measures than on a single forward pass. However, this method is slightly slower since it performs multiple forward passes instead of a single one.

\citet{coreset} argue that the uncertainty based methods for active learning in CNNs are ineffective since they query a set of images instead of a single image per iteration, causing the selection of correlated images. They propose a geometric approach as an alternative, in which a subset -- called core-set -- is selected in a way that the trained model has similar performance to when the full dataset is used.  Our method is related to this work in a way that we also use geometric information (clustering), but our goal is not to select the most informative samples for the annotation, but to maximize the number of data points that will be annotated.

\citet{ensemble} use ensembles as a way of measuring uncertainty in CNNs, by using predictions produced by multiple models. \citet{Beluch_2018_CVPR} show that ensembles outperform other methods for active learning for image classification. One problem with the method is that it requires training several models instead of one, requiring more computational resources.

\citet{berardo2015active} utilize clustering in active learning for MNIST handwritten digits classification. They first employ unsupervised training for extracting features, then use these features for clustering samples. Following, they ask the expert to annotate the closest sample to each centroid and set the same label to all samples in the cluster.

\section{Active Learning with Cluster Annotation}

\begin{figure}[ht]
  \centering
  \fontsize{30}{30}\selectfont
  \begin{subfigure}[b]{.22\textwidth}
      \includegraphics[width=\textwidth]{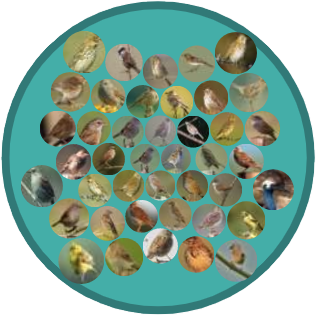}
  \end{subfigure}
  \begin{subfigure}[b]{.22\textwidth}
      \includegraphics[width=\textwidth]{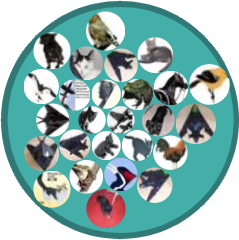}
  \end{subfigure}
  \begin{subfigure}[b]{.22\textwidth}
      \includegraphics[width=\textwidth]{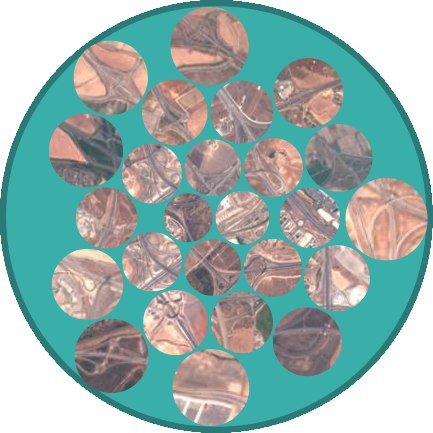}
  \end{subfigure}
  \begin{subfigure}[b]{.22\textwidth}
      \includegraphics[width=\textwidth]{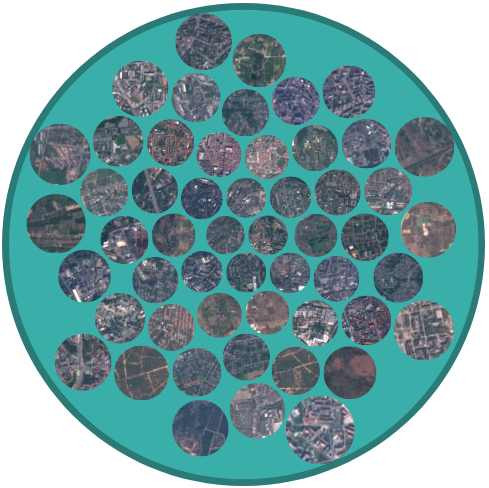}
  \end{subfigure}
  \vspace{0.3cm}
  \caption{Examples of clusters that would be presented to experts for annotation. Left to right: CIFAR-10 cluster that would be labeled as \textit{bird}; CIFAR-10 cluster that would not be annotated; EuroSAT cluster that would be labeled as \textit{highway}; EuroSAT cluster that would not be annotated. This figure is best viewed on screen.}
  \label{fig:clusters}
\end{figure}

The Active Learning with Cluster Annotation framework consists of adding the data clustering and cluster annotation steps into the standard pool-based active learning framework.

During the annotation step, the framework asks the human expert to label clusters. The framework can also ask for individual samples annotation based on some acquisition criteria (e.g., the most uncertain samples), which is the typical procedure for pool-based active learning. When annotating both clusters and individual samples, the order of annotation matters: experts can first annotate the selected individual samples and then annotate clusters, or vice-versa.

The annotation order is significant because each procedure will remove samples from the unlabeled pool. Individual samples selected for annotation will not be clustered. Likewise, samples annotated through clustering will not be considered for individual samples selection. Annotating the most informative samples first may clean the subsequent clusters, since they are more likely to end up in a wrong cluster. Clustering first may lead to a higher number of total labeled samples.

Another option is to train the model between each annotation procedure, which can increase the quality of subsequential annotations and the model performance in general. However, we did not consider this scenario since it requires two times more training steps.

The cluster annotation step is performed by presenting a visualization of samples that belong to a cluster to the expert, who can then decide to label the cluster if the majority of samples are from the same class. Otherwise, the expert does not annotate the cluster. For image classification, experts may be able to quickly skim through images inside the cluster and decide if they want to annotate it. Since the work and time required for annotating a cluster or an individual image are similar, we consider each of them as one human interaction. Figure~\ref{fig:clusters} shows examples of clusters that would be presented to experts for annotation.

The proposed framework is formally described in Appendix~\ref{app:algorithm}.

\section{Experiments}

We applied active learning with cluster annotation for image classification with two different datasets: CIFAR-10~($50\,000$ training images, $10\,000$ test images, 10 classes)~\cite{cifar10}, and EuroSAT ($27\,000$ images, 80/20 random splits, 10 classes)~\cite{eurosat}. We chose CIFAR-10 because it is widely used in image classification research and EuroSAT as it has a more specific domain and real-life applicability. CIFAR-10 classes are very related to those on ImageNet dataset (used for pretraining the models), while EuroSAT categories are unrelated with ImageNet classes.

We fine-tuned a ResNet-18~\cite{DBLP:conf/cvpr/HeZRS16} for 5 epochs for CIFAR-10, and 8 epochs for EuroSAT on every iteration. The network was pretrained on ImageNet with weights provided by PyTorch\footnote{\url{https://pytorch.org/}}. Images were resized to $224\times224$ before fed to the network. We used SGD with learning rate $1\mathrm{e}{-3}$ for CIFAR-10, and $1\mathrm{e}{-4}$ for EuroSAT, weight decay $5\mathrm{e}{-4}$, and momentum $0.9$. We extracted features from the average pooling layer (512 dimensions for ResNet-18) and used then in k-means clustering with 40 iterations (implemented by the Faiss library\footnote{\url{https://github.com/facebookresearch/faiss}}), which did not impact training time.

Networks were reset to pretrained ImageNet weights at the beginning of every iteration. Improvements in performance are due to the fact that the total number of annotated images increases after every iteration (see Figure~\ref{fig:clusters_info}). Experiments were repeated five times for statistical significance.

To simulate experts annotating clusters, we automatically assigned a label to a cluster if the modal class in the cluster corresponded to at least 80\% of the samples in it. Otherwise, the cluster was not annotated, and its images were kept in the unlabeled pool.

We experimented with five different scenarios:

\begin{enumerate}[label=\alph*)]
\item \texttt{\textbf{random}}: randomly select samples for annotation.
\item \texttt{\textbf{uncertain-only}}: use maximum entropy of softmax for selecting images for annotation;
\item \texttt{\textbf{cluster-only}}: only annotate clusters;
\item \texttt{\textbf{uncertain+cluster}}: first annotate uncertain samples, then annotate clusters;
\item \texttt{\textbf{cluster+uncertain}}: first annotate clusters, then annotate uncertain samples.
\end{enumerate}

We consider the annotation of one single image or one cluster as one human interaction. The numbers of clusters were chosen in a way that each cluster had between 50 and 100 samples on average. We used 1000 and 400 clusters per iteration for CIFAR-10 and EuroSAT, respectively, for the \texttt{cluster-only} scenario. Similarly, for the \texttt{uncertainty-only} scenario, we annotated the most 1000 and 400 uncertain samples per iteration, for CIFAR-10 and EuroSAT, respectively. For the \texttt{uncertain+cluster} and \texttt{cluster+uncertain} scenarios, we annotated 500 clusters and 500 samples, and 200 clusters and 200 samples for CIFAR-10 and EuroSAT, respectively.

\section{Results}

\begin{figure}[ht]
  \centering
  \fontsize{30}{30}\selectfont
  \begin{subfigure}[b]{1\textwidth}
      \includegraphics[width=\textwidth]{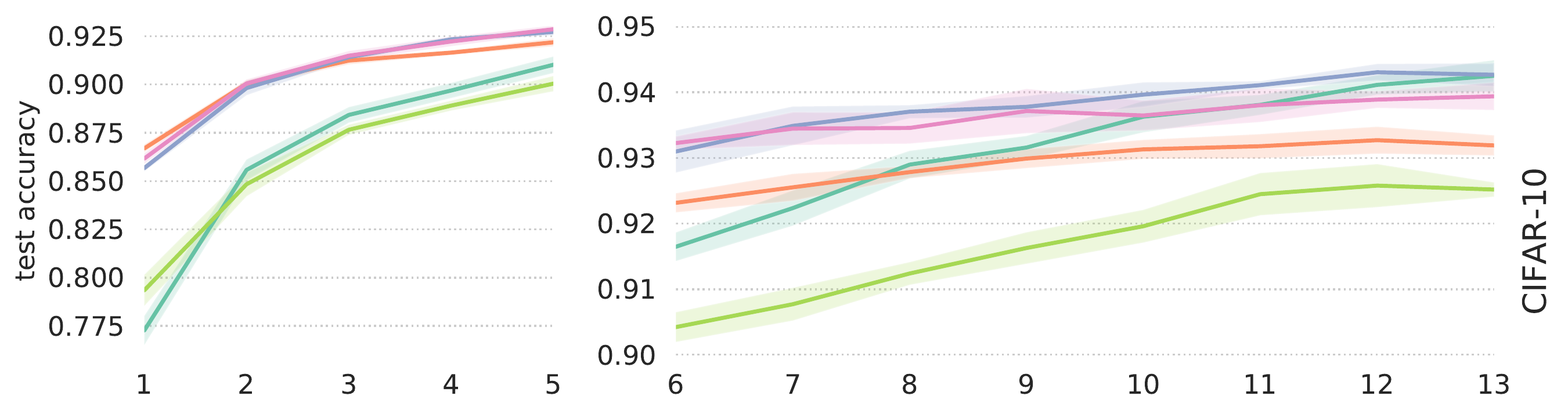}      
  \vspace{-0.8cm}
  \label{fig:cifar_test_acc}
  \end{subfigure}
  
    \begin{subfigure}[b]{1\textwidth}
      \fontsize{30}{30}\selectfont
      \includegraphics[width=\textwidth]{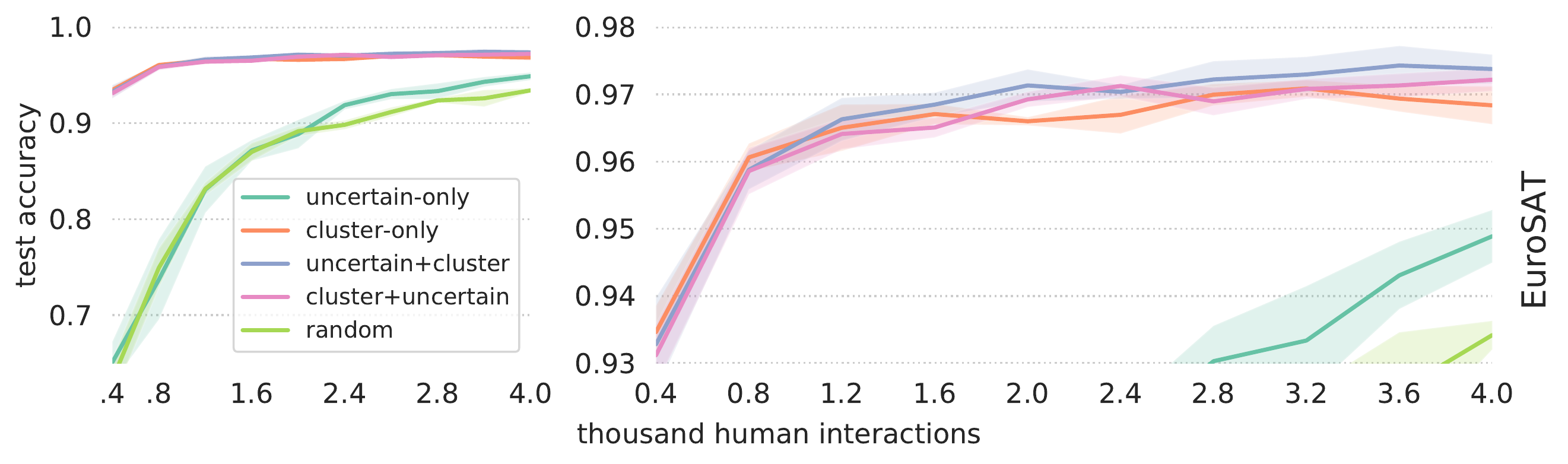}      
  \vspace{-0.8cm}
  \label{fig:eurosat_test_acc}
  \end{subfigure}
  
  \caption{Test-accuracy per thousand human interactions for CIFAR-10 (top) and EuroSAT (bottom). Each view (left and right) shows different limits in both axes for better visualization. We considered the annotation of one single image or one cluster as one human interaction. Shade areas represent standard deviations.}
  \label{fig:test_acc}
\end{figure}

\begin{figure}[ht]
  \centering
  \fontsize{30}{30}\selectfont
  \begin{subfigure}[b]{1\textwidth}
      \includegraphics[width=\textwidth]{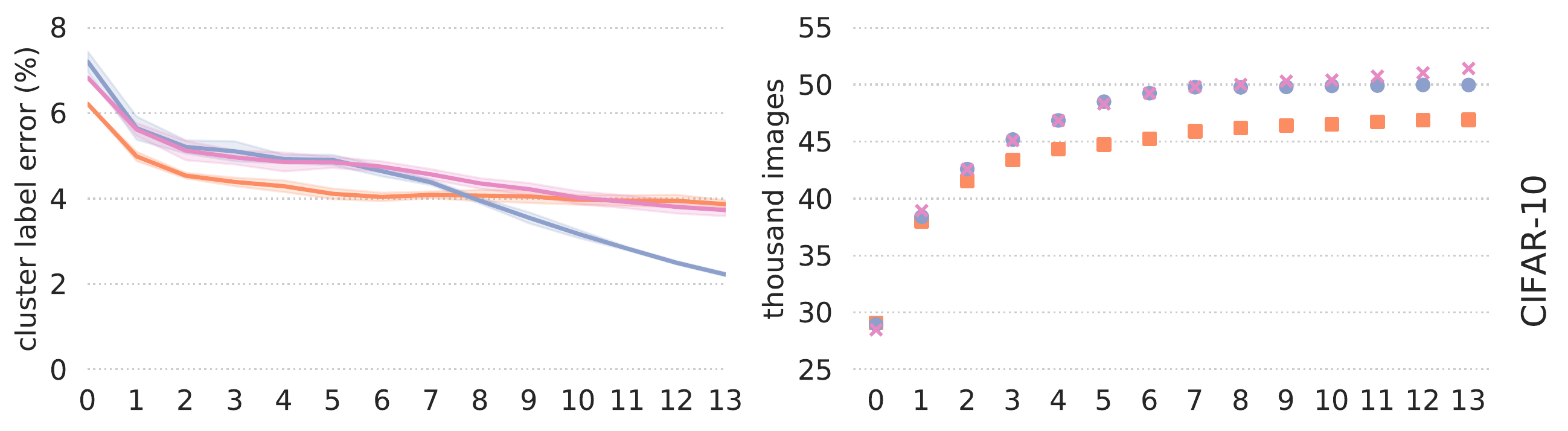}      
  \vspace{-0.8cm}
  \end{subfigure}
  
    \begin{subfigure}[b]{1\textwidth}
      \fontsize{30}{30}\selectfont
      \includegraphics[width=\textwidth]{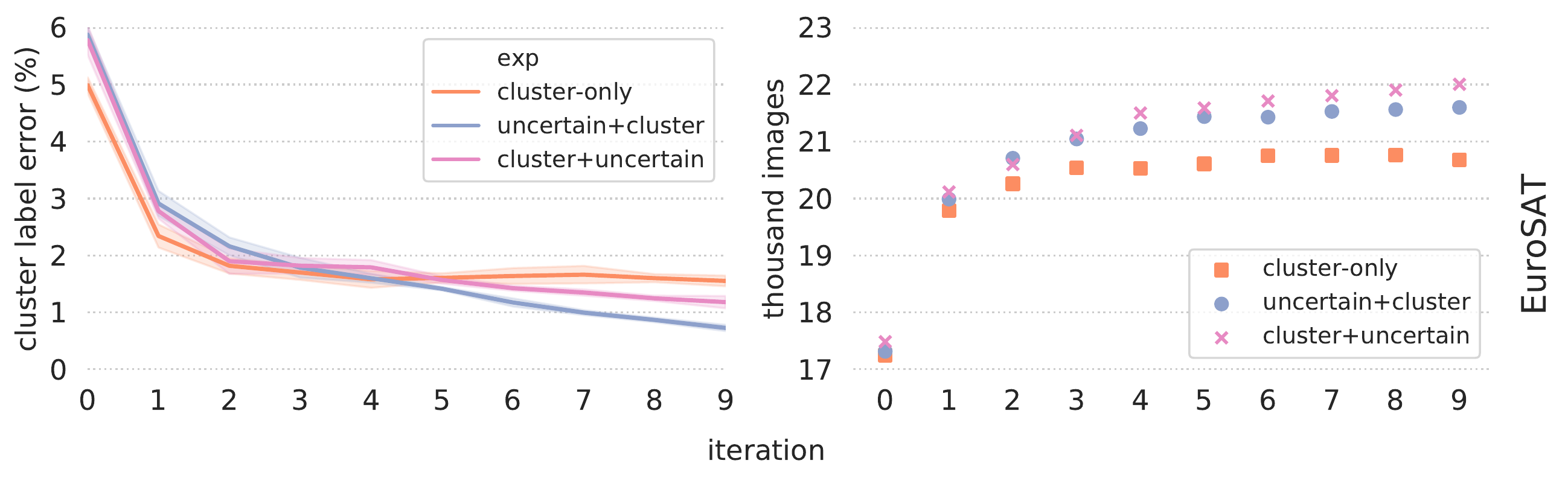}
  \end{subfigure}
  
  \caption{Label error rate for samples annotated with clustering (left), and total number of annotated images (right) per iteration. Shade areas represent standard deviations. Selecting uncertain images before clustering reduces the label error of cluster-annotated samples. Selecting uncertain samples increase the total number of annotated images.}
  \label{fig:clusters_info}
\end{figure}

Figure~\ref{fig:test_acc} shows that the test accuracy for cluster-based scenarios (\texttt{cluster-only}, \texttt{uncertain+cluster}, \texttt{cluster+uncertain}) grows much faster than scenarios without clustering (\texttt{random}, \texttt{uncertain-only}) for both datasets. However, \texttt{cluster-only} stagnates after some iterations, demonstrating the importance of also annotating samples based on uncertainty selection. This is confirmed with \texttt{uncertain+cluster} and \texttt{cluster+uncertain}, which outperform other scenarios. Scenario \texttt{uncertain+cluster} slightly surpasses \texttt{cluster+uncertain} in terms of test accuracy.

Figure~\ref{fig:clusters_info} (left) shows that the label error of cluster-annotated samples gets lower in \texttt{uncertain+cluster} than in other cluster-based scenarios, indicating that running uncertainty selection first may serve as a cluster cleaner.

Figure~\ref{fig:clusters_info} (right) shows that the total number of annotated samples increases every iteration. Scenarios with both clustering and uncertainty selection led to more annotated data than \texttt{cluster-only}, with \texttt{cluster+uncertain} having the highest values. However, this increase in annotated data does not indicate an improvement in performance when compared with \texttt{uncertain+cluster}, probably due to the higher label error rate.

Under same conditions, test accuracies for supervised training are $0.950\pm0.003$ for CIFAR-10, and $0.973\pm0.001$ for EuroSAT. With the proposed framework, CIFAR-10 has similar results from fully-supervised (50,000 images) with 9,000 (18\%) human interactions, and EuroSAT with 2,800 (13\%) human interactions (against 21,600 in supervised learning).

\section{Discussion}

We introduced active learning with cluster annotation and demonstrated that it can reduce the number of human interactions needed to train a CNN for image classification. The framework uses the same model to perform classification and to extract features for clustering; thus, not impacting training time. Moreover, by taking advantage of more data, the framework needs fewer training iterations to achieve similar results to fully-supervised learning.

The proposed framework can still be improved to achieve better results: using training techniques that are more robust to label noise; better feature extraction and clustering methods; better training conditions (e.g., different architectures, data augmentation, better hyperparameters etc.); better individual sample selection criteria.

There is also space for better cluster visualizations. Instead of displaying all images in the cluster, we could present a representative subset with fewer samples, reducing the cluster's visual complexity. \citet{berardo2015active} annotates the image that is closest to the centroid for each cluster. However, displaying only one sample per cluster may introduce too much label noise for complex datasets. Nonetheless, displaying more samples may be performed in different fashions, spanning from simply taking samples that are closest to the centroids to trying to cover a larger area of the cluster.

We believe that the framework can be applied to other domains, such as natural language processing. However, there are two main difficulties when applying it to other domains: knowledge transfer, which works well in computer vision, but is still developing in other areas; and how to present clusters for annotation, which can be trivial for images, but is much harder for other domains such as audio and text. In a future work, we would like to apply the framework to the problem of text classification.

\section*{Acknowledgements}

We gratefully acknowledge the support of NVIDIA Corporation with the donation of GPUs used for this research.

\bibliographystyle{unsrt}
\bibliography{refs}

\newcommand{\dcluster}{$D^{CL}$}
\newcommand{\dlabeled}{$D^{L}$}
\newcommand{\dunlabeled}{$D^U$}
\newcommand{\dval}{$D^V$}
\let\oldnl\nl% Store \nl in \oldnl
\newcommand{\nonl}{\renewcommand{\nl}{\let\nl\oldnl}}

\begin{appendices}
\section{Active Learning with Cluster Annotation Algorithm}
\label{app:algorithm}

\begin{algorithm}[H]
\DontPrintSemicolon
% Python Style
\SetStartEndCondition{ }{}{}%
\SetKwProg{Fn}{def}{\string:}{}
\SetKwFunction{Range}{range}%%
\SetKw{KwTo}{in}\SetKwFor{For}{for}{\string:}{}%
\SetKwIF{If}{ElseIf}{Else}{if}{:}{elif}{else:}{}%
\SetKwFor{While}{while}{:}{fintq}%
\SetKwFor{ForEach}{foreach}{:}{fintq}%
\SetKwSwitch{Switch}{Case}{Other}{switch}{:}{case}{otherwise}{endcase}
\AlgoDontDisplayBlockMarkers\SetAlgoNoEnd\SetAlgoNoLine%

\SetKwInOut{Input}{Inputs}
\Input{\texttt{T}: number of iterations.\\
\texttt{N}: number of annotation interactions per iteration.\\
\texttt{train\_dataset}: training dataset. \\
\texttt{val\_dataset}: validation dataset. \\
\texttt{annotation}: how to annotate data (\texttt{uncertain-only},\\~~~~\texttt{cluster-only},  \texttt{uncertain+cluster}, \texttt{cluster+uncertain}).\\
\textsc{SelectSamples}($dataset$, $M$): function to select the $M$ most\\~~~~informative samples from a $dataset$.\\
\textsc{Clusterize}($dataset$, $M$): function to divide a $dataset$ into $M$\\~~~~clusters.\\
}
\SetKwInOut{Variables}{Variables}
\Variables{
  \texttt{\dunlabeled}: unlabeled set. \\
  \texttt{\dlabeled}: labeled set. \\
  \texttt{\dcluster}: cluster-labeled set. \\
}
 
 \vspace{0.5cm}
 \dunlabeled $\leftarrow$ \texttt{train\_dataset}\;
 \SetKwProg{Repeat}{repeat}{ times:}{}
 \Repeat{T}{
  Initialize \texttt{model}\;
  \Switch{\texttt{annotation}}{
    \Case{uncertain-only}{
      \textsc{AnnotateUncertain}($N$)\;
    }
    \Case{cluster-only}{
      \textsc{AnnotateClusters}($N$)\;
    }
    \Case{uncertain+cluster}{
      \textsc{AnnotateUncertain}($N/2$)\;
      \textsc{AnnotateClusters}($N/2$)\;
    }
    \Case{cluster+uncertain}{
      \textsc{AnnotateClusters}($N/2$)\;
      \textsc{AnnotateUncertain}($N/2$)\;
    }
}
 Train \texttt{model} with data from \dlabeled~$\cup$ \dcluster\;
 Evaluate \texttt{model}~with data from \texttt{val\_dataset}\;
 Move all samples in \dcluster~to \dunlabeled\;
}

\vspace{0.5cm}

\SetKwProg{Fn}{Function}{:}{end}
\Fn{\textsc{AnnotateUncertain}(N)} {
 \texttt{samples} $\leftarrow$ \textsc{SelectSamples}(\dunlabeled, $N$)\;
 The expert annotates all \texttt{samples}\;
 Move the labeled samples from \dunlabeled~to \dlabeled\;
 \textbf{return}\;
}

\vspace{0.5cm}

\Fn{\textsc{AnnotateClusters}(N)} {
 \texttt{clusters} $\leftarrow$ \textsc{Clusterize}(\dunlabeled, $N$)\;
\ForEach{cluster \texttt{c} in \texttt{clusters}}{
\If{the expert believes that \texttt{c} has class consistency}{
The expert annotates \texttt{c} with label \texttt{l}\;
All samples in \texttt{c} are assigned with label \texttt{l}\;
Move all samples in \texttt{c} from \dunlabeled~to \dcluster\;
}
}
\textbf{return}\;
}
\caption{Active Learning with Cluster Annotation}
\end{algorithm}

\end{appendices}

\end{document}